\documentclass[11pt]{article}

\usepackage{graphicx}
\usepackage{verbatim}
\usepackage{amssymb}
\usepackage{amsbsy}
\usepackage{amscd}
\usepackage{amsmath}
\usepackage{amsthm}

\newcommand{\RR}{\mathbb R}

\begin{document}

\title{Asymptotic learning curve and renormalizable condition 
in statistical learning theory}

\author{Sumio Watanabe\\
P\&I Lab., Tokyo Institute of Technology\\
4259 Nagatsuta, Midoriku, Yokohama, 226-8503 Japan\\
E-mail: swatanab @ pi . titech . ac . jp}



\maketitle

\begin{abstract}
Bayes statistics and statistical physics have the common 
mathematical structure, where the log likelihood function corresponds to
the random Hamiltonian. 
Recently, it was discovered that 
the asymptotic learning curves in Bayes estimation 
are subject to a universal law, even if the log likelihood function
can not be approximated by any quadratic form. 
However, it is left unknown
what mathematical property ensures such a universal law. 
In this paper, we define a renormalizable condition of the statistical
estimation problem, and show that, under such a condition,
the asymptotic learning curves are ensured to be subject to the 
universal law, even if the true distribution is unrealizable and singular
for a statistical model. 
Also we study a nonrenormalizable case, in which the learning curves have 
the different asymptotic behaviors from the universal law. 
\end{abstract}

\section{Introduction}

In recent studies, it was pointed out that Bayes statistics and 
statistical physics have the common mathematical structure, where
the log likelihood function plays the same role as 
the random Hamiltonian, and the Bayes posterior distribution 
can be understood as the Boltzmann distribution. 
However, there are some differences
between them. In statistical learning theory,
the random Hamiltonian can not be necessarily approximated
by any quadratic form because the Hessian matrix of the log 
likelihood function can be singular \cite{NOLTA1995}. For example, 
artificial neural networks \cite{NN2001}, 
normal mixtures\cite{Yamazaki},
reduced rank regressions \cite{Aoyagi}, 
Bayes networks \cite{Rusakov}, 
binomial mixtures, Boltzmann machines, and 
hidden Markov models are singular models. 

The statistical properties of such models
have been left unknown in statistics and information science,
because it was difficult to analyze a singular likelihood function 
\cite{NOLTA1995,Hartigan}. 
Recently, new statistical learning theory has 
been established based on algebraic geometry, by which
it was proved that 
the generalization and training errors are subject to a 
universal law, even if the 
statistical model does not satisfy the regularity condition
\cite{NC2001,Drton,Cambridge2009,NN2010,IEICE2010,ASPM2010}.
However, it is not yet clarified what mathematical
properties ensure that such a universal law holds, 
therefore it is unknown the range of statistical problems
which are subject to the universal law. 

In this paper, we define a renormalizable condition 
of a statistical problem. The renormalizable condition 
requires that 
the variance function of the random Hamiltonian
is bounded by the average one. We show that, if 
a statistical problem is renormalizable, then 
the algebraic geometrical method can be successfully 
applied, resulting that the learning curves are subject to 
the universal law. Also we show that, if it is not
renormalizable, then 
the large fluctuation of the random Hamiltonian 
prevents the system from obeying to the universal law in general. 

\section{Bayes Learning Theory} 

Let $q(x)$ be a probability density function on $N$ 
dimensional real Euclidean space ${\RR}^{N}$. 
The training samples and the testing sample
are respectively defined by random variables 
$X_{1},X_{2},...,X_{n}$ and $X$, which
are independently subject to the
same probability distribution $q(x)dx$. 

A statistical model is defined as 
a probability density function $p(x|w)$ of $x\in {\RR}^{N}$ for a given 
parameter $w\in W\subset {\RR}^{d}$, where $W$ is a set of all parameters. 
In Bayes estimation, we prepare a probability density function $\varphi(w)$ on
$W$. Although $\varphi(w)$ is called a prior distribution, it 
does not necessary represent an {\it a priori} knowledge of 
the parameter, in general. 

For a given function $F(w)$ on $W$, its expectation value 
$\langle F(w) \rangle$ with respect to the
posterior distribution is defined by
\[
\langle F(w) \rangle =\frac{\displaystyle
\int F(w)\;\prod_{i=1}^{n}p(X_{i}|w)^{\beta}\;\varphi(w)dw
}{\displaystyle
\int \prod_{i=1}^{n}p(X_{i}|w)^{\beta}\;\varphi(w)dw
},
\]
where $0<\beta<\infty$ is the inverse temperature. 
The case $\beta=1$ is most important because it 
corresponds to the strict Bayes estimation. 
The Bayes predictive distribution is defined by
\[
p^{*}(x)=\langle p(x|w)\rangle.
\]
In Bayes learning theory, the following random variables 
play an important role. The Bayes generalization loss $B_{g}$,
the Bayes training loss $B_{t}$, the Gibbs generalization loss
$G_{g}$, and the Gibbs training loss $G_{t}$ are respectively
defined by 
\begin{eqnarray}\label{eq:bg}
B_{g}&=&-E_{X}[\log \langle p(X|w)\rangle ],\\
B_{t}&=&-\frac{1}{n}\sum_{i=1}^{n}\log \langle p(X_{i}|w)\rangle,\\
G_{g}&=&-\Bigl\langle E_{X}[\log p(X|w)]\Bigr\rangle,\\
G_{t}&=&-\Bigl\langle \frac{1}{n}\sum_{i=1}^{n}\log p(X_{i}|w)\Bigr\rangle,
\end{eqnarray}
where $E_{X}[\;\;]$ shows the expectation value over $X$. 
Let us introduce two random variables by 
\begin{eqnarray}
Y_{g}&=&E_{X}
\Bigl[\langle (\log p(X|w))^{2}\rangle
-\langle \log p(X|w)\rangle^{2}\Bigr],\\
Y_{t}&=&\frac{1}{n}\sum_{i=1}^{n}\Bigl\{
\Bigl\langle (\log p(X_{i}|w))^{2}\Bigr\rangle
-\Bigl\langle \log p(X_{i}|w)\Bigl\rangle^{2}\Bigr\},
\label{eq:yt}
\end{eqnarray}
where $V_{g}=nY_{g}$ and $V_{t}=nY_{t}$ are referred to
as the {\it functional variances} \cite{Cambridge2009,NN2010}.
In this paper, we study the 
expectation values of these six random variables, which are
called {\it Bayes observables}. 
The log loss function $L(w)$ and the entropy $S$ are respectively defined by
\begin{eqnarray*}
L(w)&=&-E_{X}[\log p(X|w)],\\
S&=&-E_{X}[\log q(X)].
\end{eqnarray*}
Note that $L(w)=S+D(q||p_{w})$, where $D(q||p_{w})$ is the relative entropy 
or Kullback-Leibler distance defined by 
\[
D(q||p_{w})=\int q(x)\log \frac{q(x)}{p(x|w)}dx.
\]
Therefore, always $L(w)\geq S$. Moreover, $L(w)=S$ 
if and only if $p(x|w)=q(x)$.
In this paper, 
we assume that there exists a parameter $w_{0}\in W$ 
which minimizes $L(w)$,
\[
L(w_{0})=
\min_{w\in W} L(w).
\]
Note that such $w_{0}$ is not unique in general, because the map 
$w\mapsto p(x|w)$ is not one-to-one in general. 
We assume that, for an arbitrary $w$ that satisfies $L(w)=L(w_{0})$, $p(x|w)$ is 
the same probability density function. Let $p_{0}(x)$ be such a unique probability 
density function. For simplicity, we use notation $L_{0}=-E_{X}[\log p_{0}(X)]$. 
\vskip3mm\noindent
{\bf Definition}. If 
$q(x)=p_{0}(x)$, then $q(x)$ is said to be {\it realizable} by $p(x|w)$, 
if otherwise it is said to be {\it unrealizable}. 
\vskip3mm\noindent
{\bf Definition}. If 
the set $W_{0}=\{w\in W;p_{0}(x)=p(x|w)\}$ consists of a single point $w_{0}$ and if 
the Hessian matrix $J\equiv\nabla\nabla L(w_{0})$ is strictly positive definite, then 
$q(x)$ is said to be {\it regular} for $p(x|w)$. 
If otherwise, then $q(x)$ is said to be {\it singular} for $p(x|w)$. 
\vskip3mm\noindent
Bayes learning theory was studied in realizable and regular cases
\cite{Schwarz,Levin,Amari}, realizable and singular cases
\cite{NC2001,Cambridge2009,NN2010}, and 
unrealizable and regular cases \cite{IEICE2010}. In such cases, it was
proved that there exists a universal relation between the generalization and
training errors. 
In this paper, we mainly study unrealizable and singular cases. 

\section{Generating Function of Statistical Learning}

The log density ratio function $f(x,w)$ and the log likelihood ratio function
$H_{n}(w)$ are respectively defined by
\begin{eqnarray*}
f(x,w)&=&\log\frac{p_{0}(x)}{p(x|w)}, \\
H_{n}(w)&=&\frac{1}{n}\sum_{i=1}^{n}f(X_{i},w),
\end{eqnarray*}
where $nH_{n}(w)$ is referred to as the random Hamiltonian. 
In this paper, we introduce the generating function of Bayes learning theory by 
\[
F_{n}(\alpha)=E\Bigl[-\log \int \exp(-\alpha f(X,w)-\beta n H_{n}(w))\varphi(w)dw\Bigr],
\]
where $E[\;\;]$ shows the expectation value over $X_{1},X_{2},..,X_{n}$ and $X$.
Then, by the definitions eq.(\ref{eq:bg})-eq.(\ref{eq:yt}) and by
using the fact that $\log p_{0}(x)$ is a constant function of $w$, 
it immediately follows that 
\begin{eqnarray}
E[B_{g}]&=& L_{0}+F_{n}(1)-F_{n}(0),\\
E[B_{t}]&=& L_{0}+F_{n-1}(1+\beta)-F_{n-1}(\beta), \\
E[G_{g}]&=& L_{0}+F_{n}'(0), \\
E[G_{t}]&=& L_{0}+F_{n-1}'(\beta),\\
E[Y_{g}]&=& -F_{n}''(0), \\
E[Y_{t}]&=& -F_{n-1}''(\beta).
\end{eqnarray}
These equations show that $F_{n}(\alpha)$ 
determines the behaviors of average Bayes observables \cite{NC2001,Levin,Amari}. 
In order to analyze these values, we need assumptions. 
\vskip3mm
\noindent{\bf Definition}. If 
there exists a constant $\gamma>0$ such that
\begin{eqnarray}
\lim_{n\rightarrow\infty}\;\;
\sup_{0\leq \alpha\leq 1+\beta}\;\;|F_{n}^{(3)}(\alpha)|n^{\gamma}=0,
\\
\lim_{n\rightarrow\infty}
|F_{n}'(0)-F_{n-1}'(0)|n^{\gamma}=0,
\\
\lim_{n\rightarrow\infty}
|F_{n}''(0)-F_{n-1}''(0)|n^{\gamma}=0,
\end{eqnarray}
then the generating function is said to 
satisfy the {\it conditions of learnability} with index $\gamma$.
\vskip3mm\noindent
Let us assume that the conditions of learnability are satisfied. Then, 
by using Taylor expansions of $F_{n}(\alpha)$, 
$F_{n}'(\alpha)$, and $F_{n}''(\alpha)$, 
it follows that 
\begin{eqnarray}
E[B_{g}]&=& L_{0} + F_{n}'(0)+\frac{1}{2}F_{n}''(0)+ o(\frac{1}{n^{\gamma}}), \\
E[B_{t}]&=& L_{0} + F_{n}'(0)+\frac{2\beta+1}{2}F_{n}''(0)+o(\frac{1}{n^{\gamma}}),\\
E[G_{g}]&=& L_{0}+F_{n}'(0), \\
E[G_{t}]&=& L_{0}+F_{n}'(0)+\beta F_{n}''(0)+o(\frac{1}{n^{\gamma}}),\\
E[Y_{g}]&=& -F_{n}''(0), \\
E[Y_{t}]&=& -F_{n}''(0)+o(\frac{1}{n^{\gamma}}).
\end{eqnarray}
Therefore, we obtain the {\it equations of states in statistical learning}, 
\begin{eqnarray}\label{eq:eqst1}
E[B_{g}]&=&E[B_{t}]+\beta E[Y_{t}]+o(\frac{1}{n^{\gamma}}), \\
E[G_{g}]&=&E[G_{t}]+\beta E[Y_{t}]+o(\frac{1}{n^{\gamma}}).\label{eq:eqst}
\end{eqnarray}
That is to say, if the conditions of learnability are satisfied, then 
the equations of states hold. 
Minimization of both $E[B_{g}]$ and $E[G_{g}]$ 
is one of the main purposes of statistical estimation, however, 
they need the expectation value over the testing sample $E_{X}[\;\;]$, 
hence they cannot be calculated directly from training samples. On the other hand, 
$B_{t}$, $G_{t}$, and $Y_{t}$ can 
be calculated from only training samples without any direct
information about $q(x)$. In other words, 
the equations of states show that 
$E[B_{g}]$ and $E[G_{g}]$ can be estimated from training samples,
therefore $B_{t}+\beta Y_{t}$ and $G_{t}+\beta Y_{t}$ are 
information criteria which show how appropriate the set $(p(x|w),\varphi(w))$ is. 
In fact, they are equal to AIC \cite{Akaike} 
if $q(x)$ is realizable by and regular for $p(x|w)$. 
If $q(x)$ is unrealizable by or singular for $p(x|w)$, 
then AIC is not equal to the asymptotic generalization error,
whereas $B_{t}+\beta Y_{t}$ and $G_{t}+\beta Y_{t}$ are. 
Hence they are called {\it widely applicable information criteria} (WAIC)
\cite{Cambridge2009,NN2010,ASPM2010}. 

\section{Renormalizable Case}\label{section:ren}

Let us define the renormalizability. 
\vskip3mm\noindent{\bf Definition}. 
Let $W_{\epsilon}=\{w\in W; D(p_{0}||p_{w}) \leq \epsilon\}$. 
If there exist $A>0$ and $\epsilon>0$ such that 
\[
w\in W_{\epsilon}\Longrightarrow L(w)-L_{0}\geq A\;D(p_{0}||p_{w}),
\]
then the pair $(q(x),p(x|w))$ is said to be {\it renormalizable}.
If otherwise, {\it nonrenormalizable}. 
\vskip3mm
It is easy to show that, if $q(x)$ is regular for $p(x|w)$, then $(q(x),p(x|w))$ is
renormalizable. In fact, $D(p_{0}||p_{w})$ is smaller than some quadratic
form of $w-w_{0}$ in the neighborhood of unique $w_{0}$ 
and $L(w)-L_{0}$ has a positive definite
Hessian matrix. Also, it is trivial to show that, if $q(x)$ is realizable by $p(x|w)$,
then $(q(x),p(x|w))$ is renormalizable. In fact, since $q(x)=p_{0}(x)$, 
$L(w)-L_{0}=D(p_{0}||p_{w})$. However, 
if $q(x)$ is unrealizable by and singular for $p(x|w)$,  
then $(q(x),p(x|w))$ may be renormalizable or nonrenormalizable. 

In this section, we study the renomalizable case, and show that
the conditions of learnability hold with index $\gamma=1$ and that 
the Bayes observables are subject to the universal law. 

We assume that $L(w)$ is an 
analytic function of $w\in W$ and that $w\mapsto f(x,w)$ is a function-valued
analytic function. Since $\int p_{0}(x)dx=\int p_{w}(x)dx=1$, 
\[
D(p_{0}||p_{w})=\int p_{0}(x)(f(x,w)+e^{-f(x,w)}-1)dx.
\]
There exists a constant $B>0$ such that
\[
\frac{t+e^{-t}-1}{t^{2}}\geq B \;\;\;(|t|<\epsilon).
\]
By combining this inequality with the renormalizability, it follows that 
\begin{equation}\label{eq:f2}
L(w)-L_{0}\geq AB \int p_{0}(x)f(x,w)^{2}dx.
\end{equation}
Since $L(w)-L_{0}$ is an analytic function, 
we can apply resolution of singularities \cite{Hironaka,Atiyah}
to $L(w)-L_{0}$, and obtain the following result.
There exist both a real $d$-dimensional analytic manifold ${\cal M}$ and a
real analytic
map $g:{\cal M}\rightarrow W$ such that, in each local coordinate of ${\cal M}$, 
\begin{eqnarray*}
L(g(u))-L_{0}&=&u^{2k}\equiv \prod_{j=1}^{d}u_{j}^{2k_{j}},\\
|g'(u)|\varphi(g(u))&=& b(u)u^{h}\equiv b(u) \prod_{j=1}^{d}u_{j}^{h_{j}},
\end{eqnarray*}
where $k=(k_{1},k_{2},...,k_{d})$ and $h=(h_{1},h_{2},...,h_{d})$ are multiple indeces made of
nonnegative integers, $|g'(u)|$ is the Jacobian determinant of the map $w=g(u)$, and 
$b(u)>0$. 
Then, by using eq.(\ref{eq:f2}), $f(x,g(u))^{2}$ can be divided by $u^{2k}$, 
in other words, $f(x,g(u))/u^{k}$ is a well-defined analytic function. 
In fact, if $f(x,g(u))$ can not be divided by $u^{2k}$, then 
eq.(\ref{eq:f2}) does not hold. 
Hence,
there exists a function-valued analytic function $a(x,u)$ such that 
\[
f(x,g(u))=a(x,u)u^{k}.
\]
Moreover, from $L(w)-L_{0}=E_{X}[f(X,w)]$, we have $E_{X}[a(X,u)]=u^{k}$. 
Remark that both renormalizability and resolution theorem are
necessary to prove the existence of $a(x,u)$. 
Let us define an empirical process on ${\cal M}$, 
\[
\xi_{n}(u)=\frac{1}{\sqrt{n}}\sum_{i=1}^{n}\{a(X_{i},u)-u^{k}\}. 
\]
Then the probability distribution of 
$\xi_{n}(u)$ converges to that of the gaussian process $\xi(u)$, which
is uniquely determined by its average and covariance \cite{Cambridge2009,Wvan}, 
\begin{eqnarray*}
E_{\xi}[\xi(u)]&=&0, \\
E_{\xi}[\xi(u)\xi(u')]&=&E_{X}[a(X,u)a(X,u')]-E_{X}[a(X,u)]E_{X}[a(X,u')],
\end{eqnarray*}
where $E_{\xi}[\;\;]$ shows the expectation value over the gaussian process
$\xi(u)$. Moreover, the gaussian process $\xi(u)$ can be represented by
\[
\xi(u)=\sum_{j=1}^{\infty}c_{j}(u)g_{j}
\]
where $\{g_{j}\}$ are independent random variables and each $g_{j}$ is 
subject to the standard normal distribution. Then
\[
E_{\xi}[\xi(u)\xi(u')]=\sum_{j=1}^{\infty}c_{j}(u)c_{j}(u'). 
\]
The random Hamiltonian is rewritten as 
\[
nH_{n}(g(u))=nu^{2k}-\sqrt{n}u^{k}\xi_{n}(u).
\]
To study the generatining function $F_{n}(\alpha)$, 
we need the asymptotic behavior of 
\[
Z_{n}(s)=\int f(x,w)^{s}\exp(-\beta n H_{n}(w))\varphi(w)dw,
\]
where $s\geq 0$ is a real value. For example, 
\begin{eqnarray}
F_{n}'(0)&=& E\Bigl[
\frac{Z_{n}(1)}{Z_{n}(0)}\Bigr],\label{eq:F1}\\
F_{n}''(0)&=&
-E\Bigl[\frac{Z_{n}(2)}{Z_{n}(0)}\Bigr]
+E\Bigl[\frac{Z_{n}(1)}{Z_{n}(0)}\Bigr]^{2}.\label{eq:F2}
\end{eqnarray}
Then by using the function $w=g(u)$, 
\begin{eqnarray*}
Z_{n}(s)&=& \sum_{\alpha}\int du\;a(x,u)^{s}u^{sk+h}
\exp(-\beta nu^{2k}+\beta \sqrt{n}u^{k}\xi_{n}(u))b_{\alpha}(u)\\
&=& \sum_{\alpha}\int_{0}^{\infty}dt \int du\;
\frac{1}{n}\;\delta\Bigl(\frac{t}{n}-u^{2k}\Bigr)\;
a(x,u)^{s}u^{sk+h}
\exp(-\beta t+\beta \sqrt{t}\xi_{n}(u))b_{\alpha}(u),
\end{eqnarray*}
where $\sum_{\alpha}$ shows the sum over all local coordinates and
$b_{\alpha}(u)\geq 0$ satisfies $\sum_{\alpha}b_{\alpha}(u)=b(u)$. 
By using the asymptotic expansion of the Shcwartz distribution
$\delta(t/n-u^{2k})$ for $n\rightarrow\infty$ 
\cite{NC2001,Cambridge2009,NN2010,Bernstein,Gelfand,Kashiwara,Oaku,Saito}, there exists 
a Schwartz distribution $D_{\alpha}(u)$ such that 
\[
\sum_{\alpha}\frac{1}{n}\;\delta\Bigl(\frac{t}{n}-u^{2k}\Bigr)\;u^{sk+h}\;b_{\alpha}(u)
\cong \frac{(\log n)^{m-1}}{n^{\lambda+s/2}}\;t^{\lambda-1+s/2}
\Bigl(
\sum_{\alpha^{*}}D_{\alpha^{*}}(u)\Bigr),
\]
where $\lambda>0$ is the {\it log canonical threshold} defined by 
\[
\lambda = \min_{\alpha}\min_{j=1}^{d}\Bigl(\frac{h_{j}+1}{2k_{j}}\Bigr),
\]
and $m$ is the maximum number of $j$ which attains the above minimum.
Also $\sum_{\alpha^{*}}$ shows the sum over all local coordinates 
that attain the above minimum and the support of $D_{\alpha^{*}}(u)$ 
is contained in the set $\{u\in {\cal M};L(g(u))-L_{0}=0\}$. 
Hence
\[
Z_{n}(s)
\cong \frac{(\log n)^{m-1}}{n^{\lambda+s/2}}
\Bigl(\int {\cal D}(u,t)
t^{s/2}\exp(\beta \sqrt{t}\xi(u))
\Bigr).
\]
where $\int {\cal D}(u,t)$ is defined by the integration over the manifold, 
\[
\int {\cal D}(u,t)
= \sum_{\alpha^{*}}\int_{0}^{\infty}dt \int du 
D_{\alpha^{*}}(u)\; t^{\lambda-1}\;\exp(-\beta t).
\]
Let us define
\[
\hat{Z}(q,r,s)=\int {\cal D}(u,t)\;\xi(u)^{q}\;t^{r/2}\;a(x,u)^{s}\;
\exp(\beta \sqrt{t}\xi(u)).
\]
Then
\begin{equation}\label{eq:from-to}
Z_{n}(s)\cong \frac{(\log n)^{m-1}}{n^{\lambda+s/2}}
\hat{Z}(0,s,s).
\end{equation}
Firstly, since $E_{X}[a(X,u)]=u^{k}$, 
\[
E_{X}[\hat{Z}(0,1,1)]=\hat{Z}(0,2,0).
\]
Secondly, by using the partial integration of $t$
\[
\int_{0}^{\infty}dt\;t^{\lambda}e^{-\beta t+\beta\sqrt{t}\xi(u)}
=
\frac{\lambda}{\beta}
\int_{0}^{\infty}dt\;t^{\lambda-1}e^{-\beta t+\beta\sqrt{t}\xi(u)}
+
\frac{1}{2}
\int_{0}^{\infty}dt\;t^{\lambda-1/2}\xi(u)e^{-\beta t+\beta\sqrt{t}\xi(u)},
\]
it follows that
\[
\hat{Z}(0,2,0)=\frac{\lambda}{\beta}
\hat{Z}(0,0,0)+\frac{1}{2}\hat{Z}(1,1,0).
\]
And lastly, by using the partial integration over the 
gaussian process $\xi(u)$, 
\begin{eqnarray}
E_{\xi}
\Bigl[
\frac{\hat{Z}(1,1,0)}{
\hat{Z}(0,0,0)}
\Bigr]
&=& E_{\xi}\Bigl[
\int {\cal D}(u,t)\Bigl(
\sum_{j=1}^{\infty}c_{j}(u)g_{j}
\Bigr)
\frac{t^{1/2}\exp(\beta\sqrt{t}\xi(u))}
{
\int {\cal D}(u',t')\exp(\beta\sqrt{t'}\xi(u'))
} \Bigr] \nonumber \\
&=& E_{\xi}\Bigl[
\int {\cal D}(u,t)\Bigl(
\sum_{j=1}^{\infty}c_{j}(u)\frac{\partial}{\partial g_{j}}
\Bigr)
\frac{t^{1/2}\exp(\beta\sqrt{t}\xi(u))}
{
\int {\cal D}(u',t')\exp(\beta\sqrt{t'}\xi(u'))
} \Bigr] \nonumber \\
&=& \beta E_{X}
E_{\xi}
\Bigl[
\frac{\hat{Z}(0,2,2)}{
\hat{Z}(0,0,0)}
\Bigr]
- \beta E_{X}
E_{\xi}
\Bigl[
\frac{\hat{Z}(0,1,1)}{
\hat{Z}(0,0,0)}
\Bigr]^{2}\label{eq:sf},
\end{eqnarray}
where we used $E_{\xi}[\xi(u)\xi(u')]=E_{X}[a(X,u)a(X,u')]$ 
on the set $\{u;L(g(u))-L_{0}=0\}$. 
Let us define the constant $2 \nu$ by the 
right hand side of eq.(\ref{eq:sf}), where $\nu$ is
referred to as the {\it singular fluctuation}. 
Then by using eqs.(\ref{eq:F1}),(\ref{eq:F2}),(\ref{eq:from-to}), 
\begin{eqnarray*}
F_{n}'(0)&\cong& (\frac{\lambda}{\beta}+\nu)\cdot\frac{1}{n},\\
F_{n}''(0)&\cong & -\frac{2\nu}{\beta}\cdot \frac{1}{n}.
\end{eqnarray*}
Therefore, we obtained the {\it universal law} of Bayes observables, 
\begin{eqnarray}
E[B_{g}]&\cong& L_{0}+(\frac{\lambda-\nu}{\beta}+\nu)\frac{1}{n},\\
E[B_{t}]&\cong& L_{0}+(\frac{\lambda-\nu}{\beta}-\nu)\frac{1}{n},\\
E[G_{g}]&\cong& L_{0}+(\frac{\lambda}{\beta}+\nu)\frac{1}{n}, \\
E[G_{t}]&\cong& L_{0}+(\frac{\lambda}{\beta}-\nu)\frac{1}{n},\\
E[Y_{g}]&\cong & E[Y_{t}]\cong \frac{2\nu}{\beta}\cdot\frac{1}{n}.
\end{eqnarray}
In this case, we can prove that the conditions of learnability 
with index $\gamma=1$ are satisfied by the same way as \cite{Cambridge2009,NN2010}. 
Hence, equations of states hold with $\gamma=1$.

\section{Nonrenormalizable Case}

In this section, we study a nonrenormalizable case. It is still difficult to 
clarify the general nonrenormalizable case. Hence, in this section, 
we show that there exists a simple example in which the Bayes observables do not satisfy
the universal law. 
\begin{eqnarray}
q(x,y)&=&\frac{1}{2\pi}\exp(-\frac{1}{2}(x^{2}+y^{2})),\\
p(x,y|a)&=&\frac{1}{2\pi}\exp(-\frac{1}{2}\{(x-a)^{2}+(y-\sqrt{a^{4}-a^{2}+1})^{2}\}),
\end{eqnarray}
where $a\in {\RR}^{1}$ is a parameter. 
Then the relative entropy is 
\[
D(q||p_{a})=\int q(x,y)\log\frac{q(x,y)}{p(x,y|a)}dxdy
= \frac{1}{2}(a^{4}+1).
\]
Hence $D(q||p_{a})$ is minimized at $a=0$, and 
$
L_{0}=\log(2\pi)+3/2. 
$
The Hessian is given by $\partial_{a}^{2}D(q||p_{a})|_{a=0}=0$. 
Therefore $q(x)$ is unrealizable by and singular for $p(x|a)$. 
The log density ratio function is 
\[
f(x,a)=-ax-h(a)y +\frac{a^{4}}{2},
\]
where $h(a)=\sqrt{a^{4}-a^{2}+1}-1$ is a real analytic function, and 
\[
D(p_{0}||p_{a})=\frac{a^{4}}{2}-h(a).
\]
Note that $D(p_{0}||p_{a})\cong a^{2}/2$ in the neighborhood of 
$a=0$. On the other hand, $L(a)-L_{0}=a^{4}/2$, resulting that 
$(q(x),p(x|a))$ is not renormalizable. 
The random Hamiltonian is 
\[
n H_{n}(a)=\frac{n\;a^{4}}{2}-
\sqrt{n}\;a\;\xi_{1}-\sqrt{n}\;h(a)\;\xi_{2},
\]
where 
\[
\xi_{1}=\frac{1}{\sqrt{n}}\sum_{i=1}^{n}X_{i},\;\;\;
\xi_{2}=\frac{1}{\sqrt{n}}\sum_{i=1}^{n}Y_{i}
\]
are independently subject to the standard normal distribution. 
\begin{eqnarray*}
nH_{n}(a)'&=& 2\;n\;a^{3}-
\sqrt{n}\;\xi_{1}-\sqrt{n}\;h(a)'\;\xi_{2},\\
nH_{n}(a)''&=& 6\;n\;a^{2}-\sqrt{n}\;h(a)''\;\xi_{2}.
\end{eqnarray*}
The parameter $a$ that minimizes $nH_{n}(a)'$ is denoted by $a^{*}$.
Since
\[
a^{*}=\Bigl(\frac{\xi_{1}}{kn}\Bigr)^{1/3}+o_{p}(\frac{1}{n^{1/3}}),
\]
the main order term of $nH_{n}(a)$ is given by 
\begin{eqnarray*}
nH_{n}(a)&=&\frac{1}{2}nH_{n}(a^{*})(a-a^{*})^{2}+nH_{n}(a^{*})\\
&=& \frac{1}{2}C_{n}(a-D_{n})^{2}-\frac{1}{2}C_{n}D_{n}^{2},
\end{eqnarray*}
where
\begin{eqnarray*}
C_{n}&=& 6 n (\xi_{1}/2\sqrt{n})^{2/3},\\
D_{n}&=& (\xi_{1}/2\sqrt{n})^{1/3}.
\end{eqnarray*}
Therefore, by using $E[(\xi_{1})^{\mu-1}]=2^{\mu/2}\Gamma(\mu/2)/\sqrt{2\pi}$, 
\begin{eqnarray*}
F_{n}'(0)&=& \frac{Q}{2}\cdot\frac{1}{n^{2/3}},\\
F_{n}''(0)&=&-\frac{2Q}{\beta}\cdot\frac{1}{n^{2/3}},
\end{eqnarray*}
where 
\[
Q= \frac{2^{7/6}}{\sqrt{2\pi}}\Gamma(\frac{7}{6}).
\]
The asymptotic behaviors of Bayes observables are different from
the universal law, 
\begin{eqnarray}
E[B_{g}]&\cong& L_{0}+(\frac{1}{2}-\frac{1}{\beta})\cdot\frac{Q}{n^{2/3}},\\
E[B_{t}]&\cong& L_{0}-(\frac{3}{2}+\frac{1}{\beta})\cdot\frac{Q}{n^{2/3}},\\
E[G_{g}]&\cong& L_{0}+\frac{Q}{2}\cdot\frac{1}{n^{2/3}},\\
E[G_{t}]&\cong& L_{0}-\frac{3Q}{2}\cdot\frac{1}{n^{2/3}},\\
E[Y_{g}]&\cong& E[Y_{t}]\cong\frac{2Q}{\beta}\cdot\frac{1}{n^{2/3}}.
\end{eqnarray}
Also in this case, the conditions of learnability are satisfied with index $2/3$, hence
the equations of states hold with $\gamma=2/3$, however, 
\[
E[V_{t}]=n E[W_{t}]\cong n^{1/3}
\]
does not converge to the constant. It seems that both renormalizable and nonrenormalizable 
statistical problems satisfy the more general universal law. 

\section{Discussion}

In this section, let us discuss three points, 
birational invariants, renormalizability, and
Bayes observables as random variables. 

\subsection{Birational Invariants}

In section \ref{section:ren}, we proved that, in the 
renormalizable case, the asymptotic learning curves are
determined by $\lambda$ and $\nu$, which are defined by
using resolution of singularities. 
Let us study the mathematical properties of them. 
For a given analytic function, $L(w)-L_{0}$, there exist 
infinitely many desingularization pairs $({\cal M},g)$. 
If a value defined by using $({\cal M},g)$ does not 
depend on the choice of $({\cal M},g)$, then 
it is called a {\it birational invariant}. 

Firstly, as is shown in \cite{NC2001,Cambridge2009}, 
the value $(-\lambda)$ is equal to the largest pole of
the {\it zeta function} on ${\mathbb C}$ 
obtained by the analytic continuation 
of 
\[
\zeta(z)=\int (L(w)-L_{0})^{z}\varphi(w)dw\;\;\;
(\mbox{Re}(z)>0). 
\]
Therefore, $\lambda$ is a birational invariant. 
This value is well known in algebraic geometry 
and algebraic analysis, which shows the relative relation
of the pair of two algebraic varieties $(W,W_{0})$
\cite{Hironaka,Bernstein,Kollor,Mustata,Oaku,Saito}.

Secondly, the value $\nu$ is characterized by
\[
\nu=\lim_{n\rightarrow\infty}
\frac{\beta}{2}E\Bigl[
\frac{1}{n}\sum_{i=1}^{n}\Bigl\{
\Bigl\langle (\log p(X_{i}|w))^{2}\Bigr\rangle
-\Bigl\langle \log p(X_{i}|w)\Bigl\rangle^{2}\Bigr\}
\Bigr].
\]
Hence $\nu$ is also a birational invariant. 

It was clarified by \cite{IEICE2010} that, if a true distribution is
unrealizableby and regular for a parametric model, then
\begin{eqnarray*}
\lambda&=&d/2, \\
\nu&=&\mbox{tr}(IJ^{-1})/2,
\end{eqnarray*}
where $I$ and $J$ are respectively $d\times d$ matrices defined by
\begin{eqnarray*}
I&=&E_{X}[\nabla\log p(x|w_{0})\nabla\log p(x|w_{0})],\\
J&=&\nabla^{2}L(w_{0}).
\end{eqnarray*}
For singular and realizable cases, 
$\lambda$ was calculated in \cite{Yamazaki,Aoyagi,Rusakov,NN2010},
whereas $\nu$ is unknown. 

\subsection{Renormalizability}

Let us discuss the renormalizable condition. 

Firstly, we study the renormalizable condition from
the physical point of view. In physics, a set of 
functions $\{f_{n}(x);n=1,2,...,\}$ is sometimes called 
renormalizable if there exists some rescaling transform by which 
a universal law is discovered. For example, 
if there exist both a set $(a,b)$ and a function $f^{*}(x)$ such that 
\[
\lim_{n\rightarrow\infty}
n^{a}f_{n}(n^{b}x)\rightarrow f^{*}(x),
\]
then such a system is called renormalizable. 
In this paper, we have shown that, if $(q(x),p(x|w))$ is renormalizable, then 
the Boltzmann distribution satisfies the convergence in law,
\[
\frac{
n^{\lambda}}{(\log n)^{m-1}}\;\exp(-n\beta H_{n}(g(u)))\rightarrow 
\int_{0}^{\infty} t^{\lambda-1}\exp(-n\beta t+\beta \sqrt{t}\;\xi(u))dt,
\]
when $n$ tends to infinity, where $\xi(u)$ is a gaussian process defined by
the central limit theorem of the functional space. 
If $(q(x),p(x|w))$ does not satisfy the renormalizable
condition, then such a rescaling transform does not exist in general. 
The expectation and the variance of
\[
nH_{n}(w)=\sum_{i=1}^{n}f(X_{i},w)
\]
are respectively given by 
\[
E[nH_{n}(w)]=n E_{X}[f(X,w)],\;\;\;
V[nH_{n}(w)]=n V_{X}[f(X,w)].
\]
Because $E_{X}[f(X,w)]=L(w)-L_{0}\geq 0 $ and $V_{X}[f(X,w)]\cong (1/2)D(p_{0}||p_{w})$
in the neighborhood $L(w)-L_{0}=0$, the renormalizable condition 
ensures that the fluctuation of the random Hamiltonian is bounded by
the average one.
This is the intuitive reason why the universal law
holds. 

Secondly, we study scale invariantness of renormalizablity. 
Let $f_{1}(x,w)$ and $f_{2}(x,w)$ be log likelihood ratio
functions of two different statistical problems. If 
they are renormalizable and satisfy the relations
\begin{eqnarray*}
E_{X}[f_{1}(X,w)]&=&E_{X}[f_{2}(X,w)], \\
E_{X}[f_{1}(X,w)f_{1}(X,w')]&=&E_{X}[f_{2}(X,w)f_{2}(X,w')],
\end{eqnarray*}
then they have the same birational invariants $(\lambda,\nu)$. 
In other words, the learning curves are determined only by the
average and covariance of the log density ratio function. 
It might seem that $E_{X}[f(X,w)]^{2}\propto E_{X}[f(X,w)^{2}]$,
but such a relation does not hold even in a trivial case. 
In a realizable and regular case, $a\in {\RR}^{1}$, 
\[
p(x|a)=\frac{1}{(2\pi)^{1/2}}\exp(-\frac{1}{2}(x-a)^{2})
\]
and $q(x)=p(x|0)$, then $
f(x,a)=a^{2}/2-ax$, resulting that
$E_{X}[f(X,a)]=a^{2}/2$ and $E_{X}[f(X,a)^{2}]\cong a^{2}+a^{4}/4$.
Therefore, in the neighborhood of $a=0$, both $E_{X}[f(X,a)]$ and 
$E_{X}[f(X,a)^{2}]$ are in proportion to $a^{2}$. The renormalizable
condition in this case is invariant under a scaling transform $f(X,w)
\rightarrow s f(X,w)$ for an arbitrary constant $s>0$. The renormalizable 
condition of this paper is a generalized concept of such invariantness. 

\subsection{Bayes Observables as Random Variables}

In statistical learning theory, Bayes observables are 
random variables. In this paper, we mainly studied the
expectation values of them. Note that 
the generating function $F_{n}(\alpha)$ does not have 
sufficient information about randomness of Bayes 
observables. If a true distribution is regular or realizable,
then stochastic properties of Bayes observables were 
clarified \cite{NN2010,IEICE2010}. It is a future study
to clarify the stochastic behavior of Bayes observables as random variables.

\section{Conclusion}

In this paper, we defined the renormalizable condition of
a learning system, and proved that, in the renormalizable 
case, the universal law holds. Also we showed that,
in nonrenormalizable case, the universal law does not hold
in general. It is the future study to clarify the more general
universal learning theory, which contains both renormalizable and
nonrenormalizable statistical problems. 

\vskip3mm\noindent{\bf Acknowledgment}.
This research was partially supported by the Ministry of Education,
 Science, Sports and Culture in Japan, Grant-in-Aid for Scientific
 Research 18079007.


\end{document}